\documentclass[runningheads,a4paper]{llncs}

\usepackage{amssymb}
\usepackage{graphicx}
\usepackage{algorithm2e}
\usepackage[acronym]{glossaries}
\usepackage[table]{xcolor}

\usepackage{url}
\urldef{\mailsa}\path|{alfred.hofmann, ursula.barth, ingrid.haas, frank.holzwarth,|
\urldef{\mailsb}\path|anna.kramer, leonie.kunz, christine.reiss, nicole.sator,|
\urldef{\mailsc}\path|erika.siebert-cole, peter.strasser, lncs}@springer.com|    
\newcommand{\keywords}[1]{\par\addvspace\baselineskip
\noindent\keywordname\enspace\ignorespaces#1}

\newacronym{pon}{PON}{Passive Optical Network}
\newacronym{gpon}{GPON}{Gigabyte Passive Optical Network}
\newacronym{ftth}{FTTH}{Fiber-to-the-Home}
\newacronym{fttx}{FTTX}{Fiber-to-the-X}
\newacronym{fttb}{FTTB}{Fiber-to-the-Building}
\newacronym{olt}{OLT}{Optical Line Terminals}
\newacronym{onu}{ONU}{Optical Network Units}
\newacronym{co}{CO}{Central Office}
\newacronym{jso}{JSO}{Joint Optical Splitter}
\newacronym{isp}{ISP}{Internet Service Provider}
\newacronym{sdus}{SDU}{Single Dwelling units}
\newacronym{mdus}{MDU}{Multi Dwelling units}
\newacronym{pdo}{PDO}{Optical Distribution Point}

\begin{document}

\mainmatter  

\title{Automatic Design of Telecom Networks with Genetic Algorithms}
\author{Jo\~ao Correia\inst{1} \and Gustavo Gama\inst{1} \and João Tiago Guerrinha\inst{2} \and Ricardo Cadime\inst{2} \and Pedro Antero Carvalhido\inst{2} \and Tiago Vieira\inst{2} \and Nuno Lourenço\inst{1}}

\institute{University of Coimbra, CISUC, DEI, Coimbra, Portugal \\ \texttt{jncor@dei.uc.pt, gustavogama@student.dei.uc.pt, naml@dei.uc.pt}
\and Altice Labs, Portugal\\ \texttt{ \{joao-t-guerrinha,ricardo-v-cadime,pedro-a-carvalhido\\,tiago-s-vieira\}@alticelabs.com}}


\authorrunning{Correia et al.}
\maketitle

\begin{abstract}
With the increasing demand for high-quality internet services, deploying GPON/Fiber-to-the-Home networks is one of the biggest challenges that internet providers have to deal with due to the significant investments involved. Automated network design usage becomes more critical to aid with planning the network by minimising the costs of planning and deployment. The main objective is to tackle this problem of optimisation of networks that requires taking into account multiple factors such as the equipment placement and their configuration, the optimisation of the cable routes, the optimisation of the clients' allocation and other constraints involved in the minimisation problem. An AI-based solution is proposed to automate network design, which is a task typically done manually by teams of engineers. It is a difficult task requiring significant time to complete manually. 
To alleviate this tiresome task, we proposed a Genetic Algorithm using a two-level representation to design the networks automatically. To validate the approach, we compare the quality of the generated solutions with the handmade design ones that are deployed in the real world. The results show that our method can save costs and time in finding suitable and better solutions than existing ones, indicating its potential as a support design tool of solutions for GPON/Fiber-to-the-Home networks. In concrete, in the two scenarios where we validate our proposal, our approach can cut costs by 31\% and by 52.2\%, respectively, when compared with existing handmade ones, showcasing and validating the potential of the proposed approach.

\end{abstract}

\section{Introduction}
In the last few years, the demand for high-quality internet services increased rapidly due to the growth in the usage of bandwidth-intensive applications such as Cloud-based Services, Video and Audio Streaming, and gaming. This pushes \gls{isp} to consider faster technologies to answer the increasing demand, such as optical fiber, i.e. \gls{fttx}. \gls{fttx} refers to a wide range of deployment configurations for different scenarios in the last mile stage of telecommunications networks. The most common scenarios are the \gls{ftth} connecting directly to the client's residence, and \gls{fttb} reaching buildings directly. One way of providing this type of service is through a \gls{gpon} based solution. \gls{gpon} is a point-to-multipoint (P2MP) network which enables a single fiber system to serve multiple customer premises with a splitting process. This technology has many advantages compared to other types of networks ( e.g., copper-based networks (DSL)), which have been the standard solution for many years. First, it allowed for faster data transmission and increased range, with optical fiber cables able to cover distances up to 20km. Secondly, it allows for boosted security. Thirdly, if there is the need to upgrade the network, it is possible to change only the endpoints and leave the fiber infrastructure intact. Furthermore, lastly, it is cost effective. 

To deploy a GPON/FTTH network, there are many challenges that the \gls{isp} has to consider, involving significant investments. Deployment of the GPON/FFTH networks involves building new infrastructures (e.g., duct digging, laying new cables to connect to the homes) or updating and maintaining an existing one. As such, it is crucial to optimise the design of the network to reduce high costs. Considering that the source of difficulties while deploying a network is engineering or technical aspects of the design, proper design of the solution can lead to savings in the range of $30\%$ \cite{FTTHHandbook}. 

However, to create a cost-effective GPON/FTTH, engineers have to consider several factors, such as the splitter position and its maximum ratio, the headend position, the optical distribution point position, the routes to consider, the  maximum distance, the number of homes ones has to serve and the optical budget. Considering all these factors, it is a complex task requiring considerable time and effort. Therefore, automation of the process can minimise the effort required for the design process from days to a few hours or minutes whilst optimising the design in what concerns costs. Typically this process design process has several phases, from the physical infrastructure, passing from the design planning of the GPON/FFTH networks to the network itself. This goes without saying that there are multiple stages where automation can help save time and effort and be valuable for the telecom network company responsible for its deployment. Although multiple phases of the design process occur in the context of this work, we are concerned with the design of a network at the equipment level, assuming that the physical part is already handled.

Thus, this paper proposes a bio-inspired approach to the automatic design of GPON/FTTH networks. The automatic design of the network is framed as a combinatorial optimisation problem that we aim to solve using Evolutionary Computation. We designed a Genetic Algorithm (GA) where each individual has a two-level representation that maps into an existing physical infrastructure. The first level corresponds to a binary list defining the points of optimal distribution (PDO), and the second list indicates the connections between a PDO and the clients. We validate the approach by employing the proposed approach to optimise real-world networks and compare the automatically designed solutions with the existing ones, manually designed by engineering teams and currently deployed and functioning. The results show that for each benchmarked network, the GA is able to design networks with inferior costs to the existing handmade ones. In concrete, when compared to the handmade solutions, our approach can discover solutions that reduce the costs by 31\% and  52.2\% for the two scenarios considered. 

The remainder of the article is as follows. In section \ref{sec:background}, we cover the background and definitions of the problem and the related work. Section \ref{sec:approach} presents the approach for the automatic design of networks. Afterwards, in Section \ref{sec:expsetup}, we present the experimental setup and in Section \ref{sec:expresults} the results. In Section \ref{sec:conclusions}, overall conclusions are drawn, and future work is presented. 

\section{Background}
\label{sec:background}

A simple GPON network is composed of three components: the \gls{olt} that represents the shared network equipment of the \gls{isp}, the different \gls{onu}, which represents the equipment on the client side (the modem) and the splitter.

Splitter Level and Ratio There is a single feeder optical fibre from the OLT that connects the splitter and splits the signal on multiple fibres with a $1:N$ ratio that will link to the ONUs, with $N$ being the number of output ports. To optimally design a GPON/FTTH, this splitter ratio and the level problem must be solved. Multiple configurations can be used. For example, in a \textit{centralised approach}, the designer can consider a splitting ratio of $1:32$, which means that this splitter design can serve up to 32 homes (ONUs).
In a \textit{cascading splitting approach}, we can have different splitters connecting each other before reaching the user side \cite{AlgoPonCapexMin}. This approach with multiple levels may use different ratios for the splitters. For example, a $1:4$ ratio for the first splitter level is followed by a $1:8$ for the second.
The choice of approach depends on the region to serve. A centralised approach might adapt more to urban environments and crowded areas as it allows more flexibility and lowers costs. On the other hand, a cascaded approach is often used in scenarios where it is required to cover an extended area, for example, in Long-Reach Passive Optical Network \cite{cascaded}. 

The locations of the network equipment are crucial in the design of networks due to the high costs of the materials and the civil work involved. It is essential to consider the different equipment factors like the splitter's location, the location of the different ONUs (the clients), the headend position, the optical distribution point position (i.e. a street cabinet along the route used to manage interconnections, integrate fibre splicing, can also contain splitters and other components).
When planning a cost-effective network design, all the equipment must be placed optimally to minimise the amount of cable used in the network and the overall cost.

If we are dealing with buildings as homes (an FTTB scenario), we might have to consider the demand of that location. It might require, for example, a demand of $50$ ports for all the homes in the building, while in a simple house, we require only one. Thus, the cabinet containing the splitters and other components should be placed directly on the location of the building.

Another important aspect to consider is that the optical power budget has to be respected regardless of the choice of topology and design considerations. The optical power budget refers to the light required to transmit signals over a fibre connection within a certain distance. There are multiple sources of budget loss (in dB). The splitter can generate energy loss when dividing the signal, called the splitter loss (table 1). It can also happen with other components of the network, like connectors. However, most of the loss is made up by the cable length, around 0.35 dB per km. Thus, it emphasises the importance of reducing the cable required to run the GPON. By minimising the cable's distance, we improve the network's efficiency (reduced loss).

Considering the aspects mentioned above, the problem of automatically designing a GPON network can be summarised as finding the most effective locations for the different types of equipment, the split ratios of the splitters for the different levels, and the optical power budget; while reducing the financial costs.

\subsection{Related Work}

The problem of designing fibre networks is often tackled with meta-heuristics techniques. In \cite{DOPNGA}, an application of a Genetic Algorithm combined with Graph theory to design PONs for different network topologies (bus, tree and ring) is proposed for a multiple-level network (i.e., with multiple splitters levels). The main objective was to minimise the cost of implementation of the network with a focus on optimising the material needed. Like the number of splitters, their geographical location, their split ratio and the cable length. 
For that purpose, the authors' solution for the GA is based on the paths between the OLT (the shared equipment in the central office) to each ONU (the homes) obtained with the help of graph theory algorithms. Another similar work is performed in \cite{OPONP}, where a GA is applied to solve the PON optimisation problem with multiple constraints. Once again, a multiple-level splitter design was considered. A single individual is represented by a double string (where the order of the elements matter), one string for the splitters equipment location (primary and secondary level) and ratio and another for the ONUs locations. The size of those strings depends on the number of ONUs to satisfy and the available material/splitters. This kind of representation allows the authors to satisfy certain constraints automatically. The author also did a comparison work between the GA approach and an ILP approach. Both seem to find exact solutions for smaller networks. However, as the instance of the problem increases, the ILP solution needs to catch up in terms of execution time for an acceptable solution where the GA prevails, even though it is not certain that the optimal global solution is found.

A more recent work proposed by \cite{DiasGA}, also detailed an application of the GA in combination with Graph theory to solve the problem of the design of PONs. Four different topologies were considered, using different configurations of splitters, including both centralised and cascaded approaches. The representation choice for the GA consisted of a binary representation to find the locations to place the equipment amongst a set of potential candidate spots. 
Their system was tested in real networks from Brazil, where the authors compared the application of the different topologies for different maps. Overall it depends on the type of map. Dense and non-dense maps were considered. For example, a cascaded approach might be more adapted in scenarios with fewer clients to reduce cable usage. The optical budget is also considered in the solutions to check the validity of the networks. 

Other meta-heuristic approaches have also been used to address this network design problem. For example, in \cite{EnAnt}, a solution using Ant Colony Optimization (ACO) is used to design GPON/FTTH for a greenfield (i.e., deployment of a network to a fresh new site) with the input requirements of the potential locations for the materials (splitters and cable distribution locations). Their method allows them to approach near-optimal solutions. The authors compared with exact solutions solvers (simplex algorithms), and the difference in solution quality is less than $1\%$, with the advantage of a much better execution time for more significant instances ($50-90\%$ faster). Additional steps were required to achieve this result, like softening the constraints and post-optimisation. In \cite{AntCol}, the same authors propose a similar approach with ACO with extra meta-heuristics, but this time for a problem evolving additional equipment (aggregating equipment/drop closures).

In addition, in the work of Ali et al. \cite{GLS}, a Local Search approach was introduced. A Local Search procedure modifies the candidate solutions locally until no improvement is detected (or other stop criteria). In this work, those modifications can be simply the assignment of a client to new network equipment or swapping the connection (link) between two clients. As for the guided aspect, a penalty function is used when the algorithm is stuck in a local optimum. The objective function will be changed (i.e., the weights of the features present in the current solution will be modified). This approach is compared to Simulated-annealing one and seems to outperform the latter significantly. 

Finally, there are also applications of AI in other subproblems of telecommunications networks. The work of Javier Mata\cite{AIPON} proposes a survey of the vast applications of AI for telecommunication problems. Some of the referenced works are worth mentioning, for example, Morais et al. \cite{GAtopdesignSurvivable}, proposes a GA algorithm to deal with topology design considering network survivability. Network survivability refers to the capacity of the network to operate under the presence of equipment failures. Once more, ILP approaches are also used to benchmark their approach.


\section{Proposed Approach}
\label{sec:approach}

We developed a system based on Genetic Algorithms to tackel the problem of the design of networks. As seen in Section \ref{sec:background}, some design considerations must be made to obtain a robust solution for the problem. This section will overview the system developed and outline the principal algorithmic design choices considered. During the evolutionary process, the GA has to deal with the following constraints and challenges: (i) find the placement of the network equipment, precisely the optical distribution point positions (PDOs), to connect the client's homes to the network. The PDOs are capacitated, i.e. they have limited ports available; (ii) the routes to consider to minimize the resources needed to cover the totality of clients' homes, thus, the solution must consider that the routes that are "buried" (more expensive) should be avoided whenever it is possible; (iii) the demand of each client must be considered (1 unit of demand for fibre corresponds to 1 port allocated to the client); (iv) if we have big buildings such as apartments, the placement of the PDO is expected to be on-site, i.e. closest network equipment node, for "on-field" practicality, reducing the amount of drop cable in this case despite the demand of the location becoming higher; (v) the solution must find the optimal drop cable links based on the PDOs existing ports (vi) there are constraints related to maximum distance for the drop cables (e.g. limit the amount of cable that can be used to link one client to a PDO) (vii) constraints for the distribution cables, i.e., to restrict the network size and respect the optical budget; (viii) the design is impacted by weights, in the form of required equipment costs of a given solution; (ix) the constraints related to intersections of drop cables, i.e. a drop cable cannot intersect with a distribution cable.
In the next sections we detailed aspects related to the characteristics of the employed GA-based solution.

\subsection{Representation}

In our approach, each individual in the population is composed of two arrays: (i) a binary representation, where each index corresponds to a network equipment candidate to the placement of a PDO, where if the value is one, the PDO is placed on that equipment; (ii) an integer based representation used to manage the drop cables links for the capacitated sub-problem related to the limited ports of the PDOs. Each index corresponds to a different client, and the values within correspond to the associated PDO (node id) for that specific client's home. If a house is not assigned to a PDO, the value is -1. This array depends on the first binary representation, where modifying the first might impact the second.

\subsection{Variation operators}
Our variation operators are applied to the binary representation. The mutation modifies the population of individuals randomly by changing one gene. For this work we employed bit flip mutation which consists of simply modifying a bit from the first representation array, i.e., a 1 can become a 0 with a fixed probability and vice-versa.
    
    


Similarly, the crossover operator allows for variation in the population at each generation by combining the genetic material of two distinct individuals (the parents) and obtaining a new solution. The function of the crossover operator is the uniform crossover.

In the evolutionary system proposed, the variation operators modify  ``representation 1'' which can require an update of ``representation 2'' since we are potentially modifying the placement of a PDO, and some homes might need to be reinspected to check their links. Both those operators are applied for a selected pool of individuals (the parents) based on fitness scores.

\subsection{Local Search}

Concerning the sub-problem related to the capacitated problem of the PDOs mentioned earlier, there are two options for the step of connecting each home to a PDO: (i) closest to the home method; (ii) perform local search with Hill Climbing to optimise the distribution. For our specific case, we end up mixing both. The algorithm starts by performing option (i), and if we end up with homes not allocated to their closest PDO, we store them for option (ii). A computation of the possible set of PDO nodes for each home allocated to their second (or more) closest PDO is required and a cache can be used to avoid repetitive computations. Then, the swap procedure can start by swapping drop cables with clients' homes from that set of PDOs. If the cost is reduced, the modified solution is kept.


    


It is important to note that every swap has to respect the constraints in limit drop cable range and ports for the PDOs (for both the current home inspected and the other client home swapped). 

Another point worth mentioning is that we only enter this Local Search sometimes, which would be computationally costly. The algorithm only considers the Local Search step if homes are not allocated to their closest PDO, which in theory, only happens in more dense zones of the graph.

\subsection{Fitness function}

Another critical aspect of the GA is the fitness function used to evaluate each solution's performance. The evolutionary system proposed evaluates the population initially and at every generation cycle. It represents the financial cost of building the network, where the objective is to minimise this cost.

The fitness function requires the graph, the individual to be evaluated, the precalculated costs for the drop cable and the different input weights (business rules) provided by the user. Three main materials greatly impact the cost of the solution returned: (i) Drop cable cost $C_{Drop}$; (ii)Distribution cable cost $C_{Dist}$; (iii) PDO costs $C_{Pdo}$ which should depend on the maximum ports available. 

The first step consists in computing the amount of drop cable needed, $N_{Drop}$, for the current solution (this can be done by accessing the cache containing the drop cable distance costs computed in the initial phase of the system). A simple iteration through the second representation (integer one) is sufficient to perform this step. It is important to note that a factor $10$ is applied to the financial cost of the drop cable in the case of the buildings (MDUs) to guide the GA to place the PDOs near the buildings (if there is any on the map). 

The second step focuses on the solution's needed distribution cable, $N_{Dist}$. This is the network's cable between the starting point and the different PDOs (total of $N_{Pdo}$). This step needs more computational power to perform since it is required to compute the shortest path between the PDO to the starting point. Dijkstra's algorithm is applied in this scenario, and the results are stored in a dynamic cache so other calls to this fitness function can eventually reuse them. This storage step is essential when dealing with more extensive networks and dense graphs to speed up the process. 

The third step multiplies the input weights by the resources needed (cables and PDOs) with the following formula for the cost of the materials $C_{Mat}$ : 
\begin{equation}
C_{Mat} = C_{Drop} * N_{Drop} +  C_{Dist} * N_{Dist} +  C_{Pdo}*N_{Pdo} 
\end{equation}
Finally, extra penalties can be applied to this cost $C_{Mat}$ in case of broken constraints. In this case, the system may cover all the clients. So, every solution that does not have all the clients connected to the network suffers a constant penalty $P$; the constant $P$ is fixed to the cost of a PDO (chosen by trial and error). This penalty is based on the number of homes missing to be allocated $H_{Missing}$. This value will be zero if all houses are covered and no penalty is added (minimisation). The fitness ($f$) result returned is the following:
\begin{equation}
f = C_{Mat} + H_{Missing} * P 
\end{equation}

\section{Experimental Setup}
\label{sec:expsetup}

\begin{table}[t]
\centering
\caption{Aspects of the graphs for the two maps that compose the experiment.}
\label{tab:my-table-datasets} 

\begin{tabular}{|c|c|c|c|c|c|}
\hline
\textbf{\begin{tabular}[c]{@{}c@{}}Data set\\  (map)\end{tabular}} & \textbf{\begin{tabular}[c]{@{}c@{}}Number of \\ locations \\ (nodes)\end{tabular}} & \textbf{\begin{tabular}[c]{@{}c@{}}Number of \\ routes \\ (edges)\end{tabular}} & \textbf{\begin{tabular}[c]{@{}c@{}}Number of \\ valid network \\ equipment \\ nodes\end{tabular}} & \textbf{\begin{tabular}[c]{@{}c@{}}Total clients:\\ Houses(SDUs) \\ \& Buildings \\ (MDUs)\end{tabular}} & \textbf{\begin{tabular}[c]{@{}c@{}}Sum of the \\ distance of \\ all routes\end{tabular}} \\ \hline
\textit{\textbf{Map 1 }} & 188 & 81 & 80 & 103 \& 5 & 2.11 km \\ \hline
\textit{\textbf{Map 2}} & 222 & 166 & 133 & 76 \&  11 & 4.76 km \\ \hline
\end{tabular}
\end{table}

This section defines the experimental setup used to validate the proposed approach to the automatic design of networks. Our experiments aim to automatically find a solution for each dataset and compare it to the existing real-world handmade solution. We start by describing the details of the maps used, followed by the business rules and restrictions that the telecommunications partner imposed. Lastly, we present the parameters used in the Genetic Algorithm.

We evaluate our approach in two different scenarios from New York and New Jersey. Table \ref{tab:my-table-datasets} details the available maps for which we have a corresponding handmade solution. In this table, the number of location nodes (column 2) is sometimes superior to the sum of the number of valid network equipment nodes (column 4) and the total client nodes (column 5). This occurs because some of the equipment nodes got filtered since they were not part of the main connected component containing the starting point, i.e., isolated nodes.

%

The business rules comprise the materials costs and the constraints related to the problem. They have a direct impact on the result returned by the algorithm. In this set of experiments, we used a configuration aimed at fewer PDOs, less distribution cable, and more drop cable. The configuration is pre-set with costs for several items that act on weights in the evaluation phase. Ultimately encourage the algorithm to converge to a solution with the aspects mentioned earlier. Each of those scenarios depends on a fixed set of weights representing the material costs (input provided by the user). Table \ref{tab:my-table-weights} displays the experiments' weights. Note that the parameters can be adjusted depending on the restrictions. It is worth noting that these are the weights plugged into the Genetic Algorithm's fitness function.

\begin{table}
\centering
\caption{Sets of experimental weights for the materials.}
\label{tab:my-table-weights}
\begin{tabular}{|l|l|}
\hline
\textbf{Material} & \textbf{Cost} \\ \hline
\textbf{Cost PDO} & 300\$ unit \\ \hline
\textbf{\begin{tabular}[c]{@{}l@{}}Cost drop cable \\ per meter\end{tabular}} & 2\$ \\ \hline
\textbf{\begin{tabular}[c]{@{}l@{}}Cost distribution cable\\ per meter\end{tabular}} & 5\$\\ \hline
\end{tabular}

\end{table}

As for the constraints, they refer to everything that might limit the search space to find a solution. Each scenario is also limited by the constraints defined in Table \ref{tab:my-table-constraints}. There are two types of constraints to consider. The ones concerning the PDOs: maximum number of ports per PDO and also the margin to leave for each PDO for future expansion of the network. For example, if a PDO with a limit of 12 ports is chosen with a margin of 10\% to leave open, only 11 ports are operational in practice. We also have restrictions on the maximum distance allowed for the cables, like the maximum overall distance of the network (from the OLT to a client) and the limits imposed on the drop cables. For the drop cables, the designs are proposed for an 85 meters limit, but it can vary depending on the data set. Our approach is able to deal with all the mentioned inputs, i.e. costs and constraints, and are adjustable and chosen by the user that interacts with our approach. From an experimental point of view, they are the inputs for our approach.

\begin{table}
\caption{Experimental constraints.}
\label{tab:my-table-constraints}
\centering
\begin{tabular}{|l|l|}
\hline
\textbf{Constraint parameter} & \textbf{Value} \\ \hline
\textit{\textbf{Limit ports per PDO}} & 12\\ \hline
\textit{\textbf{\begin{tabular}[c]{@{}l@{}}Margin ports open per PDO\end{tabular}}} & 10\% \\ \hline
\textit{\textbf{\begin{tabular}[c]{@{}l@{}}Limit drop cable\end{tabular}}} & 85m (range: 50-100) \\ \hline
\textit{\textbf{\begin{tabular}[c]{@{}l@{}}Maximum range of the network\end{tabular}}} & 20 Km \\ \hline
\end{tabular}

\end{table}


\begin{table}
\centering
\caption{Genetic Algorithm parameters.}
\label{tab:my-tableGA}
\begin{tabular}{|c|c|}
\hline
\textbf{Parameter}                                                                                   & \textbf{Value}                                                                 \\ \hline
Population size                                                                    & 100                                                                            \\ \hline
Number of generations                                                              & 100                                                                            \\ \hline
Mutation                                                                           & \begin{tabular}[c]{@{}c@{}}Bitflip\end{tabular} \\ \hline
\begin{tabular}[c]{@{}c@{}}Mutation rate\end{tabular}            & \begin{tabular}[c]{@{}c@{}}$2/len(genotype)$\end{tabular}                    \\ \hline
Crossover                                                                          & \begin{tabular}[c]{@{}c@{}}Uniform\\ (0.5 per gene)\end{tabular}         \\ \hline
\begin{tabular}[c]{@{}c@{}}Crossover rate\end{tabular}           & 0.85                                                                            \\ \hline
Tournament selection                                                               & 5                                                                              \\ \hline
\begin{tabular}[c]{@{}c@{}}Survivors selection\\ (elitism percentage)\end{tabular} & 10                                                                             \\ \hline          
\end{tabular}

\end{table}

Table \ref{tab:my-tableGA} presents the configuration used in the GA for each evolutionary run. The experimental results that we are going to present are averages of 10 evolutionary runs. All the experiments were conducted in a computer with the following hardware specifications: Intel(R) Core(TM) i7-8700K CPU @ 3.70GHz; RAM: 16GB.


\section{Experimental Results}
\label{sec:expresults}

In this section, we present the results obtained by our approach when designing networks for two maps. We analyse and compare the obtained networks with the handmade ones, considering the following metrics: number of PDOs, drop and distribution cable used. All the results for our approach are means of 30 independent runs. 

\begin{figure}[]
\centering
\centerline{\includegraphics[width=.8\linewidth]{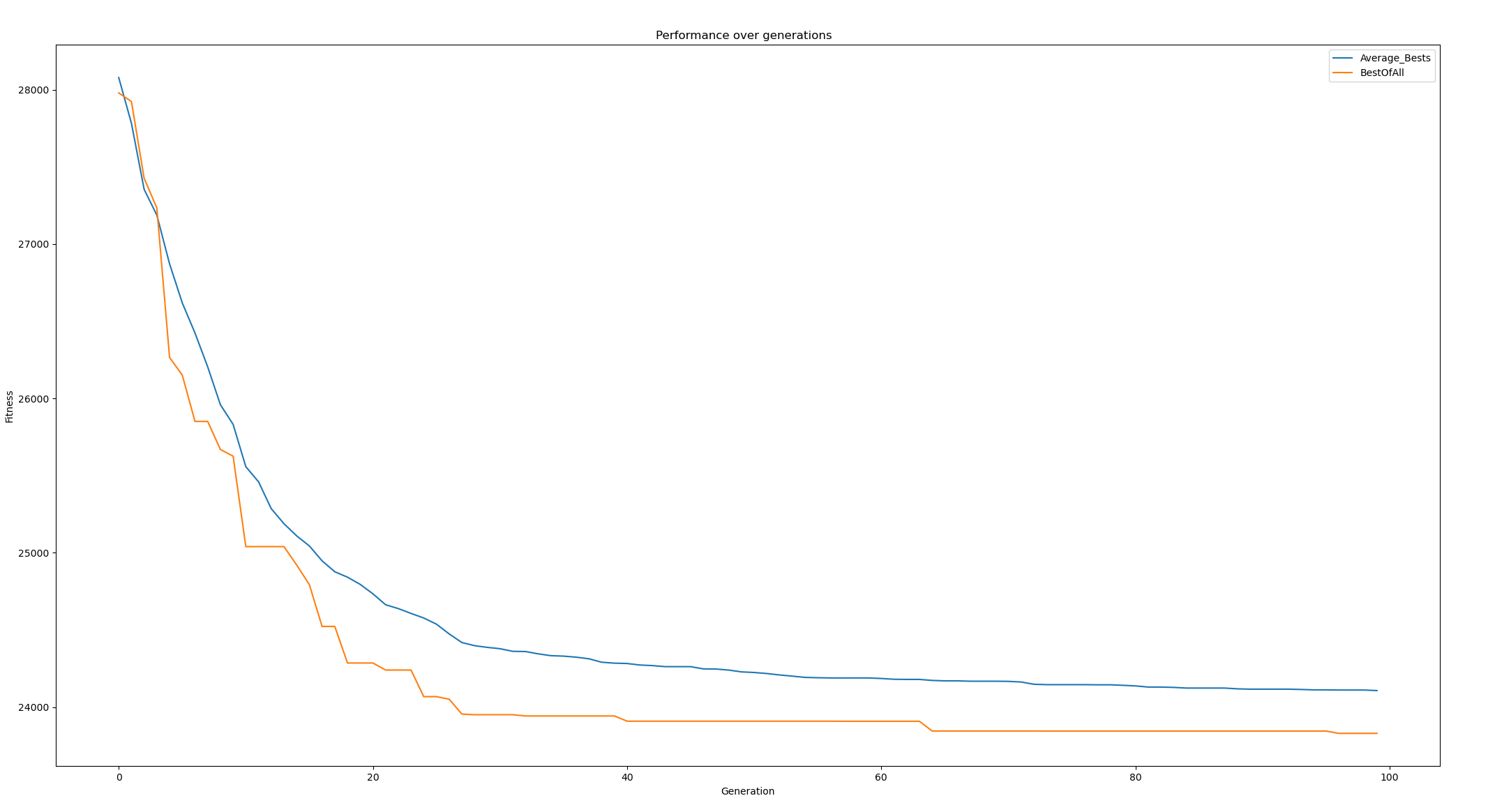}}
\caption{Fitness over the generations for the dataset Map 1.}
\label{figNY10D0170_4}
\end{figure}

Fig. \ref{figNY10D0170_4} presents the avarage and best across the 100 generations over 10 runs. Looking at the results depicted in the plot, one can see that fitness rapidly decreases in the first 25 generations. After this point, the GA continues to improve the quality of the solutions but at a slower pace. These results show that one does not need to wait long to obtain good-quality results. 

Table \ref{tab:statsExp1_1} summarises the results for the metrics. The second column shows the values for the handmade solution metrics, and the next column shows the results for the best, worst, median and means solutions obtained by the GA. When comparing the best GA solutions, we can see that the GA solutions reduce the number of PDOs (16). In contrast, it increases the distances for the drop cables.

We can also notice that the fitness score for the worst solutions increases considerably compared to the best solution (the best has a score of $23 908$ and the worst solution has a score of $204 545$). This high score is due to some penalties in the fitness function of the GA, more specifically in the case where some homes are not allocated to the network. A high constant penalty based on the homes left to be allocated is applied to guide the evolutionary algorithm towards complete coverage of the homes (which corresponds to a penetration rate of $100$\%).

The second column of the table (Handmade) contains the statistics for the handmade real-world deployed solution. When comparing it to the solutions generated by the GA, in terms of fitness score, the GA outperforms the handmade design when considering the best solution (and the median). This is because way too many PDOs are installed in the handmade solution. Every network equipment (like a pole or pedestal) has an installation point (PDO), which could be more optimal.

\begin{table}
\caption{Comparison between the handmade and the GA solutions for the Map 1.}
\label{tab:statsExp1_1}
\begin{tabular}{rr|rrrr|}
\cline{3-6}
                                                       &                   & \multicolumn{4}{c|}{\textbf{GA}}                                                                                                    \\ \hline
\multicolumn{1}{|l|}{\textbf{Metrics}}                 & \textbf{Handmade} & \multicolumn{1}{l|}{\textbf{Best}} & \multicolumn{1}{l|}{\textbf{Worst}} & \multicolumn{1}{r|}{\textbf{Median}} & \textbf{Mean}     \\ \hline
\multicolumn{1}{|l|}{Number of PDOs}                   & 62             & \multicolumn{1}{r|}{17}         & \multicolumn{1}{r|}{22}          & \multicolumn{1}{r|}{20}           & 18.80 ($\pm$0.92) \\ \hline
\multicolumn{1}{|l|}{Drop cable used (km)}             & 2.75              & \multicolumn{1}{r|}{4.18}          & \multicolumn{1}{r|}{3.76}           & \multicolumn{1}{r|}{4.07}            & 4.15 ($\pm$0.12)  \\ \hline
\multicolumn{1}{|l|}{Distribution cable used (km)}     & 2.08              & \multicolumn{1}{r|}{1.74}          & \multicolumn{1}{r|}{1.77}           & \multicolumn{1}{r|}{1.74}            & 1.70 ($\pm$0.03)    \\ \hline
\multicolumn{1}{|l|}{Fitness Score}                    & 34505          & \multicolumn{1}{r|}{23908}      & \multicolumn{1}{r|}{204545}      & \multicolumn{1}{r|}{24561}        & 38893($\pm$5832)  \\ \hline
\multicolumn{1}{|l|}{Fitness Difference (GA-Handmade)} & \cellcolor{black!25}                 & \multicolumn{1}{r|}{-10597}        & \multicolumn{1}{r|}{170040}         & \multicolumn{1}{r|}{-9944}           & -4388             \\ \hline
\end{tabular}
\end{table}

One of the requirements for the PDOs was the number of ports available and the margin of ports to leave open. Table \ref{tab:portsused_dataset1} shows the results for the PDO nodes with a building link. Those buildings (MDUs) can have a higher demand (i.e., more ports required) than a simple home (SDU) which is usually 1.
We can see that even when multiple homes and buildings are connected to the same PDO, there is always a margin-left of ports as it never reaches the maximum limit. This allows for future network expansion in case new homes are built in those areas, for example, or another existing home changes its demand. 

As for the types of splitters and cables, we needed the following extra equipment in Map $1$: $2$ splitters; $2$ cables 16FO and; $3$ cables 32FO. The demand for this network is 121 fibres. To serve that, we need at least $2$ splitters of type $1:64$. As for the cables, different types are needed based on the branches that come out of the starting point node. For example, a branch that requires $47$ fibres requires a distribution cable of 32FO and another of 16FO (which results in 48 fibres, with 1 additional that is unused but still can be helpful for future expansion).

Regarding execution time, it took $30.5$ seconds to do a single run to design this first network. This is advantageous compared to the handmade solution, which in theory, could take several hours for a network of this size. Overall, the total cost of the best-optimised version is $31\%$ cheaper than the handmade counterpart.

\begin{table}
\centering
\caption{Amount of ports used in the PDOs that contain a connection to a building (MDU) for Map 1.}
\label{tab:portsused_dataset1}
\begin{tabular}{|c|c|c|} 
\hline
\begin{tabular}[c]{@{}c@{}}\textbf{PDO node id}\\\textbf{(that contains a}\\\textbf{~MDU link)}\end{tabular} & \begin{tabular}[c]{@{}c@{}}\textbf{Ports used}\\\textbf{(out of 12 max.)}\end{tabular} & \begin{tabular}[c]{@{}c@{}}\textbf{Amount of }\\\textbf{unique clients~served}\\\textbf{(distinct }\\\textbf{SDUs/MDUs nodes)}\end{tabular} \\ 
\hline
\textbf{\textit{0}} & 6 & 5 \\ 
\hline
\textbf{\textit{9}} & 9 & 4 \\ 
\hline
\textbf{\textit{24}} & 9 & 8 \\ 
\hline
\textbf{\textit{65}} & 7 & 5 \\ 
\hline
\textbf{\textit{76}} & 11 & 5 \\
\hline
\end{tabular}

\end{table}

The second map studied is a map from New Jersey (Map 2), slightly bigger than Map 1, with more nodes and edges. This map provides a more challenging scenario since it has multiple possible paths. On Map 1, the number of possible paths was limited to reach a node on a particular branch, i.e., there was only one single path. 
Due to space constraints, we focus only on comparing the handmade and the GA solutions in terms of the metrics. Looking at the results presented in Table \ref{tab:statsExp1_2}, one can see that they are in line with the previous ones. The GA creates solutions with a small number of PDOs at the expense of increasing the distances of the drop cable. Regarding the computational time, despite being a larger map, it took, on average, 37s to design a whole network. Compared to the days it took to design the handmade solution, the computational overhead is negligible. The best solution found by the GA represents savings of 52.2\% when comparing with the cost of to the handmade.

\begin{table}
\caption{Comparison between the handmade and the GA solutions for the Map 2}
\label{tab:statsExp1_2}
\begin{tabular}{lr|rrrr|}
\cline{3-6}
                                                       & \multicolumn{1}{l|}{}                  & \multicolumn{4}{c|}{\textbf{GA}}                                                                                                                     \\ \hline
\multicolumn{1}{|l|}{\textbf{Metrics}}                 & \multicolumn{1}{l|}{\textbf{Handmade}} & \multicolumn{1}{l|}{\textbf{Best}} & \multicolumn{1}{l|}{\textbf{Worst}} & \multicolumn{1}{l|}{\textbf{Median}} & \multicolumn{1}{l|}{\textbf{Mean}} \\ \hline
\multicolumn{1}{|l|}{Number of PDOs}                   & 27                                     & \multicolumn{1}{r|}{14}            & \multicolumn{1}{r|}{14}             & \multicolumn{1}{r|}{15}              & 16 ($\pm$0.5)                  \\ \hline
\multicolumn{1}{|l|}{Drop cable used (km)}             & 3.53                                   & \multicolumn{1}{r|}{3.09}          & \multicolumn{1}{r|}{3.29}           & \multicolumn{1}{r|}{2.90}            & 3.01 ($\pm$0.01)                   \\ \hline
\multicolumn{1}{|l|}{Distribution cable used (km)}     & 2.76                                   & \multicolumn{1}{r|}{1.65}          & \multicolumn{1}{r|}{1.57}           & \multicolumn{1}{r|}{1.76}            & 1.71($\pm$0.05)                    \\ \hline
\multicolumn{1}{|l|}{Fitness Score}                    & 50216                                  & \multicolumn{1}{r|}{24828}         & \multicolumn{1}{r|}{115013}         & \multicolumn{1}{r|}{25274}           & 26635($\pm$1247)                   \\ \hline
\multicolumn{1}{|l|}{Fitness Difference (GA-Handmade)} & -                                      & \multicolumn{1}{r|}{-25388}        & \multicolumn{1}{r|}{64797}         & \multicolumn{1}{r|}{-24942}          & -23581                              \\ \hline
\end{tabular}

\end{table}

\begin{figure}
\begin{center}
\begin{tabular}{cc}

\includegraphics[width=\linewidth]{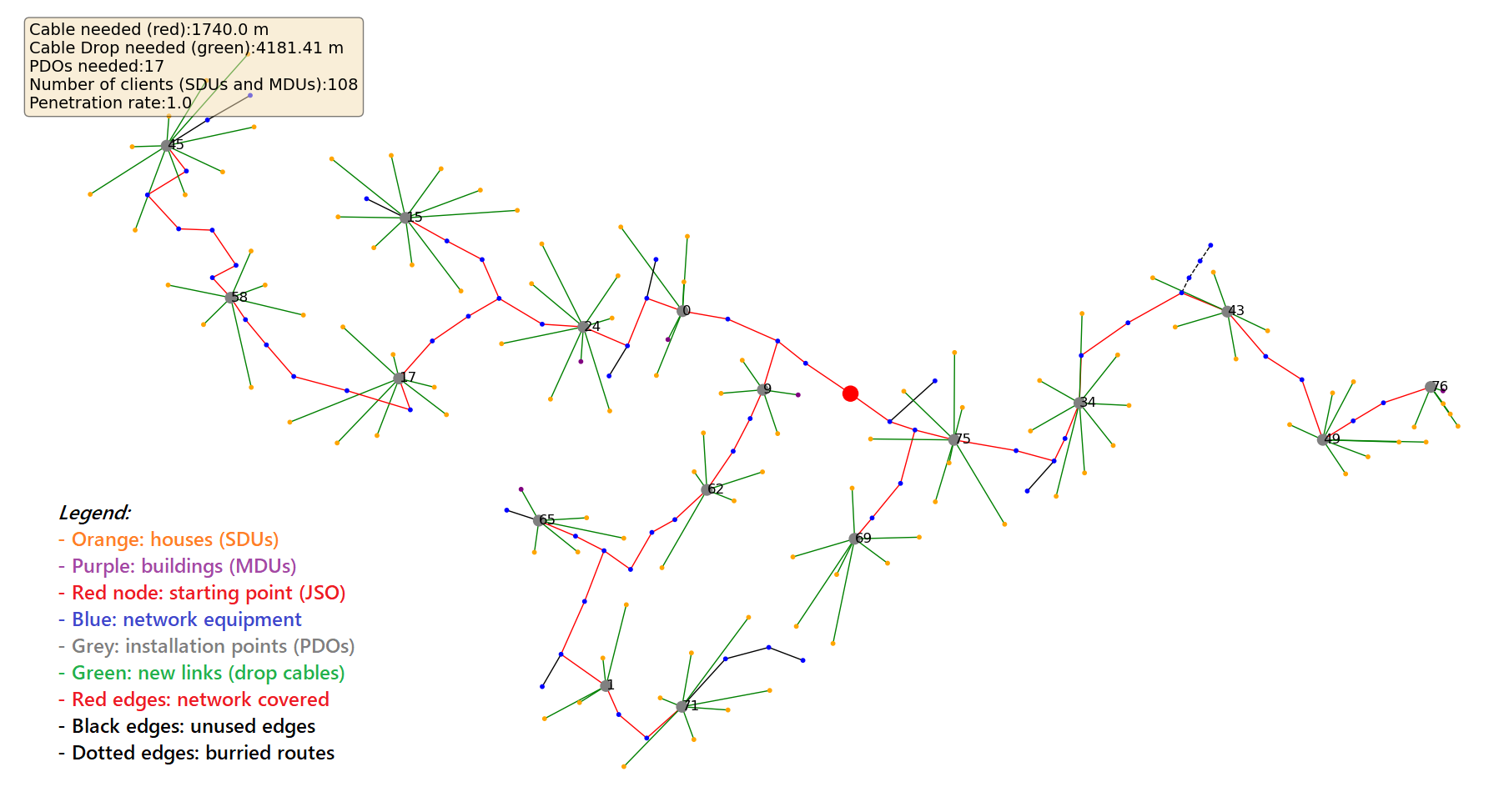}\\
\includegraphics[width=\linewidth]{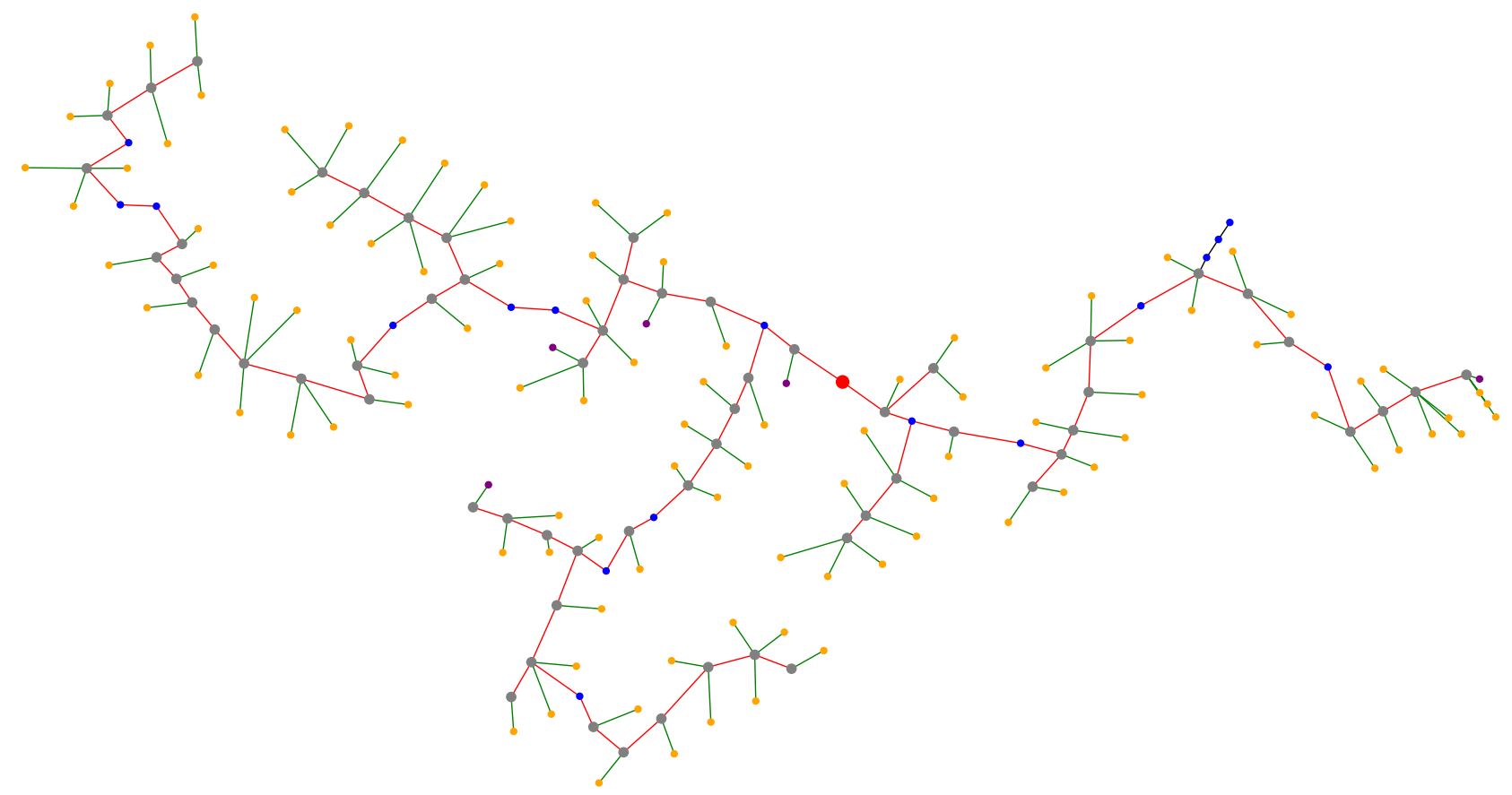}

\end{tabular}
\end{center}\vspace{-1pt}
\caption{Visualisation of the solutions for Map 1. On the top we have the best solution of the proposed approach, on the bottom we have the existing real-world handmade and deployed one. }\label{fig:surveygraphsmap1}\vspace{-0pt}
\end{figure}

\begin{figure}
\begin{center}
\begin{tabular}{cc}

\includegraphics[width=\linewidth]{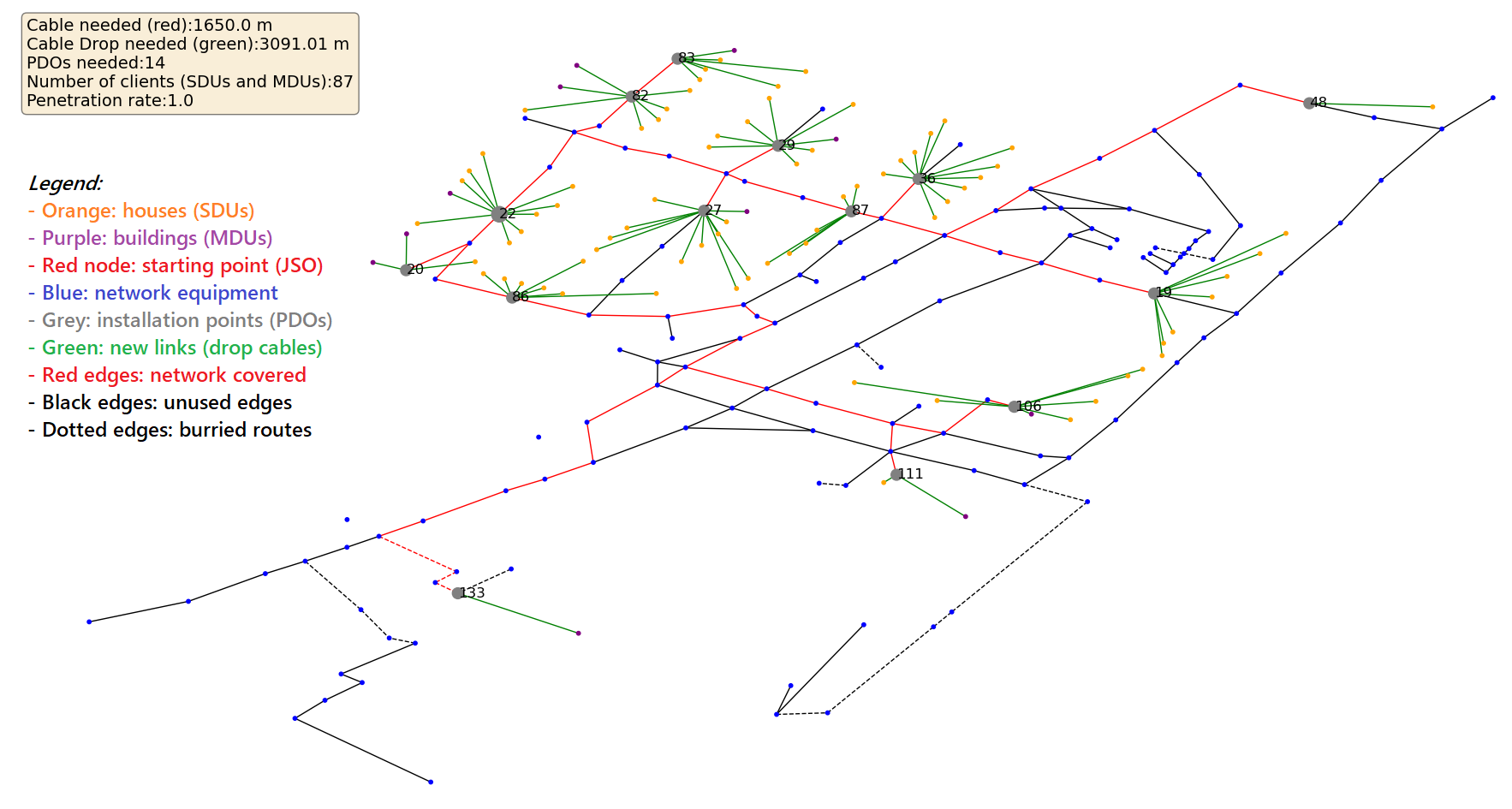}\\
\includegraphics[width=\linewidth]{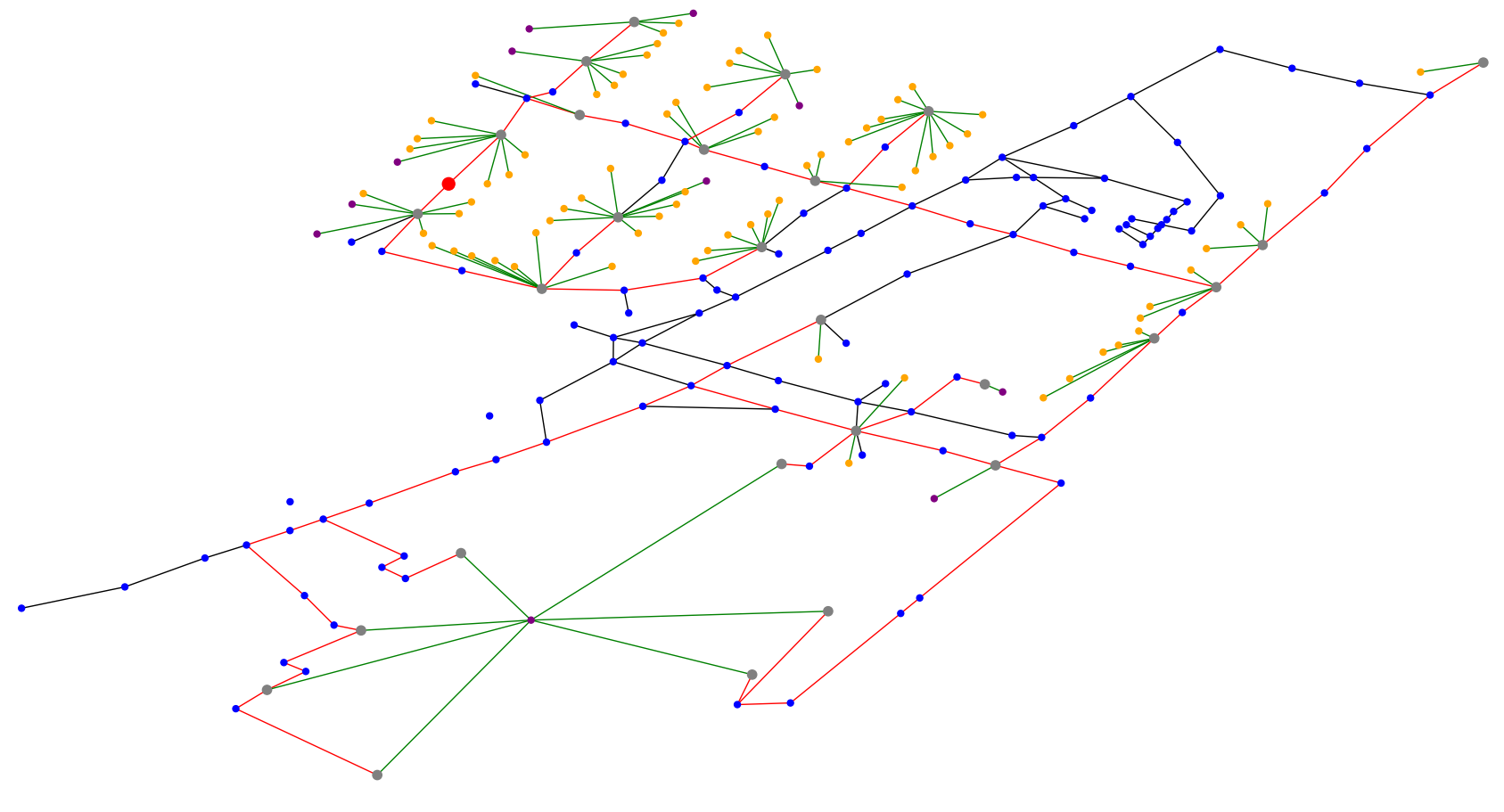}

\end{tabular}
\end{center}\vspace{-1pt}
\caption{Visualisation of the solutions for Map 2. On the top we have the best solution of the proposed approach, on the bottom we have the existing real-world handmade and deployed one. }\label{fig:surveygraphsmap2}\vspace{-0pt}
\end{figure}

In Figures \ref{fig:surveygraphsmap1} and \ref{fig:surveygraphsmap2}  we can observe the visual design solutions. In terms of solutions, they are distinct and present alternative solutions for engineers which can later be updated and enhanced. The tendency of the GA solutions is to explore and save on equipment and cut the costs of new equipment. Due to the nature of the fitness function, the outcome makes sense. We can observe that most PDOs have multiple collections when contrasting with the manually found ones. In general, the solutions found do not compromise the parameters set by the telecom company and end up giving a startup optimised and cost-saving solutions, minimising cable length, and number of PDOs and making use of the max drop cable constraint to save on multiple and short connections between clients and PDO.

\section{Conclusion}
\label{sec:conclusions}

Over the years, the demand for high-quality internet services increased rapidly due to the growth in the usage of bandwidth-intensive applications such as Cloud-based Services, Video and Audio Streaming, and gaming. This forces telecommunications companies to improve their network services to satisfy their clients. However, to create effective networks, the companies must consider several factors such as equipment cost, distance, the number of homes one has to serve and the optical budget. Considering all these factors, it is a complex task that requires time and human resources to do it. Therefore, by automating the design of these networks, we can minimise the effort required for the design process from days to a few hours or minutes whilst optimising the design in what concerns costs. 

In this work, we propose an approach based on Genetic Algorithms to design telecommunications networks automatically. In concrete, we propose a representation based on two levels. The first level corresponds to a binary list defining the points of optimal distribution, and thae second list indicates to which point a client is connected. The approach is validated by comparing the results obtained against two handmade real-world deployed solutions, using maps from New York and New Jersey. For each map, a comparison with a handmade solution was performed in order to assess the quality of the automatic design networks.  

Overall, the GA can discover better solutions than real-world existing ones that were planned and handmade by teams of engineers. For the scenarios considered, our proposed solution can reduce the costs by $31\%$ and $52.2\%$. Another important aspect of our approach is obtaining good-quality solutions in a relatively short computational time. Usually, the design of a new handmade solution can take up to multiple hours or even days of man-hours, depending on the complexity of the network. In contrast, the proposed system generates solutions in seconds, even for larger maps.

\section*{Acknowledgements}
This work is partially funded by the project POWER (grant number POCI-01-0247-FEDER-070365), co-financed by the European Regional Development Fund (FEDER), through Portugal 2020 (PT2020), and by the Competitiveness and Internationalization Operational Programme (COMPETE 2020) and by the FCT - Foundation for Science and Technology, I.P./MCTES through national funds (PIDDAC), within the scope of CISUC R\&D Unit - UIDB/00326/2020 or project code UIDP/00326/2020*

\bibliographystyle{splncs04}
\bibliography{adtnga}

\end{document}